# Automatisation de l'extraction des trajectoires 4D à partir d'enregistrements vidéo

Jean-François Villeforceix

Institut national de l'information géographique et forestière,

73 avenue de Paris, 94160 Saint-Mandé, France

jean-francois.villeforceix@ign.fr

**Résumé**

*Le Bureau d'Enquêtes et d'Analyses pour la Sécurité de l'Aviation Civile (BEA) est amené à analyser des vidéos d'accidents – issues de caméras embarquées ou au sol – impliquant tout type d'aéronef. Jusqu'alors cette analyse était manuelle et chronophage. L'objet de l'étude est donc d'identifier les cas d'application de la photogrammétrie et d'automatiser l'extraction de trajectoires 4D à partir de ces vidéos. En tenant compte de toutes les configurations de vols potentielles, des algorithmes photogrammétriques sont développés sur la base du logiciel MicMac de l'IGN et éprouvés lors de tests sur le terrain. Les résultats de ces traitements automatisés doivent se substituer aux données de vols issues d'enregistreurs type FDR[1] ou CVR[2] parfois absentes. Parmi les informations qui intéressent le BEA figurent : la position tridimensionnelle avec la composante temporelle associée, les orientations des trois axes de l'avion (angles de navigation tangage, roulis et lacet) et les vitesses moyennes (dont la vitesse ascensionnelle).*

**Mots Clef**

Géoréférencement, avion, trajectographie 4D, MicMac, calibration, aérotriangulation

**Abstract**

*The Bureau d'Enquêtes et d'Analyses pour la Sécurité de l'Aviation Civile (BEA) is led to analyze videos of accidents from on-board or earthly cameras involving all kinds of aircraft. Until today this analysis was manual and time consuming. Therefore the purpose of the study is to identify the photogrammetric use cases and to automate the extraction of 4D trajectories from these videos. Based on all potential flight configurations, photogrammetric algorithms are developed in the environment of IGN's software MicMac and tested with field trials. The results of these automated processes have to supersede missing flight data from FDR[1] or CVR[2] recorders. BEA is interested by flight information like: the 3D position with time component, the orientations of aircraft axes (navigation angles yaw, pitch and roll) and average speeds (including rate of climb).*

## 1 Introduction

Le BEA [1] est l'organisme public français responsable des enquêtes de sécurité concernant l'aviation civile. Son champ d'action concerne tous les types d'aéronefs tels que les avions, hélicoptères, planeurs, ULM, ballons, autogires, etc. dans les domaines du transport aérien public, de l'aviation générale (aviation de loisir) et du travail aérien. Lorsqu'il est chargé d'une enquête, le BEA doit déterminer le contexte et les circonstances de l'accident ou de l'incident. Pour ce faire, les appareils sont équipés d'enregistreurs de vol que les enquêteurs récupèrent pour analyse. Cependant en ce qui concerne l'aviation générale, l'enregistrement de la trace GPS n'est pas systématique alors que c'est un maillon essentiel dans la compréhension d'un accident.

L'analyse d'enregistrements vidéo, désignée par le néologisme vidéogrammétrie [2], constitue l'une des alternatives à ce manque. Qu'elle soit issue d'une *action cam* embarquée sur l'aéronef ou d'un smartphone filmant la scène au sol, la vidéo est une source d'images en recouvrement permettant une exploitation stéréoscopique de résolution variable (480p WVGA jusqu'à 4K UHD). Elle permet en théorie de dérouler un processus photogrammétrique classique à quelques ajustements près. Parmi ceux-ci figure la calibration de la caméra [3] ou du téléphone [4] qui, selon le fabricant et les conditions d'acquisition, peut s'avérer complexe. D'autres étapes s'y ajoutent comme la recherche de points d'appui terrain ou encore le ratio d'image(s) par seconde à considérer. Il en résulte alors deux cas distincts : soit la caméra est embarquée dans l'aéronef à géoréférencer, soit elle filme la scène d'accident de l'extérieur. Il s'agit donc respectivement de relever la caméra dans l'avion par rapport au terrain ou de posi-tionner un objet qui apparaît sur les images.

---

[1] *Flight Data Recorder*, enregistreurs de paramètres
[2] *Cockpit Voice Recorder*, enregistreurs phoniques

A ces deux axes d'études vient s'ajouter une configuration duale impliquant la caméra et ce qu'elle filme. C'est le cas par exemple lors de la collision de deux aéronefs dont l'un embarque la caméra qui filme l'autre aéronef impliqué. Il faut alors reconstituer la trajectoire de la caméra (et ainsi de l'avion à bord duquel elle est installée) puis positionner le second aéronef relativement à celle-ci. Cette dernière étape a fait l'objet d'un développement spécifique.

Au total ce sont donc trois configurations photogrammétriques qui sont à l'étude : le relèvement d'une caméra mobile et le positionnement d'un objet mobile sur la vidéo avec ses deux variantes – caméra fixe ou mobile. MicMac, la librairie photogrammétrique open-source développée à l'IGN, propose un certain nombre de fonctionnalités répondant aux besoins exprimés. Elles sont détaillées dans l'état de l'art qui suit. Cependant celles-ci sont parfois encore en développement ou inadaptées aux images issues de vidéo. Par ailleurs le positionnement d'un objet mobile est à élaborer puisque la caméra peut ne pas être à bord de l'appareil accidenté (vidéosurveillance) ou bien filmer un objet qui interagit avec l'avion lors de l'accident (débris, autre aéronef).

## 2 État de l'art

Les accidents d'avions sont des événements imprévisibles souvent capturés par des caméras grand public telles que celles qu'intègrent les téléphones portables. Par ailleurs la voltige, le parapente, le parachutisme, etc. sont des loisirs à sensation pour la pratique desquels les amateurs utilisent régulièrement des caméras d'action comme les GoPro [5]. Ces deux types de caméras totalement différents engendrent des difficultés au niveau de l'initialisation de la calibration. En effet certains fabricants ne communiquent pas les caractéristiques complètes de leurs appareils tandis que le processus photogrammétrique requiert certaines métadonnées EXIF de base comme la focale (éventuellement focale équivalente 35mm) ou les dimensions du capteur. C'est pourquoi il a fallu trouver le moyen de mesurer ces paramètres. L'angle de champ de la caméra est la variable photographique adéquate pour cette opération.

Par ailleurs ces grandeurs optiques sont calculées finement par les traitements itératifs de MicMac. Mais pour cela il faut des images sur lesquelles la caméra peut être calibrée [6]. Deux facteurs influent sur la calibration : le recouvrement et la résolution. L'enjeu du recouvrement est d'extraire le juste ratio d'images à la seconde afin de limiter les temps de traitements tout en conservant une redondance d'information à un seuil élevé. Quant à la résolution, l'enquêteur est tributaire du format de la vidéo. Enfin le terrain doit être bien réparti dans le champ de vue de la caméra et contrasté pour que MicMac détecte un nombre suffisant de points d'intérêt.

En ce qui concerne les méthodes de positionnement sur images, c'est la caméra qui dicte la stratégie à adopter. Si la caméra est fixe, il faudra procéder image par image avec une calibration exogène. MicMac dispose d'une fonction qui permet un relèvement de caméra par image à condition de disposer d'une pré-calibration. Elle est fondée sur un calcul de pose par relèvement dans l'espace à partir de trois correspondances. Et si la caméra ne peut pas être pré-calibrée, une variante permet de s'auto-calibrer sur chaque image en résolvant les équations dites « à 11 paramètres » de la formule d'image avec des saisies images de points terrain connus. En revanche si la caméra est mobile, il est possible de reconstituer des couples stéréoscopiques.

Pour ce faire, une fonction spécifique de traitements vidéo est nécessaire. Celle utilisée travaille avec la librairie de codecs avconv compatible Unix. Elle contient un algorithme d'extraction d'images d'une vidéo qui travaille sur deux aspects complémentaires : d'une part la netteté des images qui régule le nombre de points d'intérêts détectés et appariés ; d'autre part le recouvrement inter-images adaptatif en fonction de la vitesse de défilement du sol sur la vidéo [7]. Ce dernier aspect est une étape délicate du processus puisqu'un sous-échantillonnage peut mettre en échec la mise en place relative du jeu de données et un sur-échantillonnage provoque une augmentation exponentielle des temps de calculs.

Comme évoqué en introduction, le principal axe de recherche et de développement de cette étude concerne l'algorithme de positionnement des objets qui entrent dans le champ de la caméra. Que cette dernière soit fixe ou mobile, le principe est le même. Il faut connaître respectivement la calibration et l'orientation au préalable pour accéder d'une part à la direction de faisceau corrigée de la distorsion et d'autre part à la distance entre la caméra et l'objet.

Par ailleurs l'orientation de la caméra doit être transformée en angles de navigation exploitables pour le BEA. MicMac calcule des matrices de rotation entre le repère terrain et les sommets de prise de vue. Or ce repère caméra est photogrammétrique. Il convient donc d'appliquer une rotation 3D par rapport à chaque axe. En outre si la caméra est décalée par rapport aux axes principaux de l'avion, une seconde rotation est nécessaire afin de corriger cette différence.

Le noyau dur de la chaîne est par conséquent issu de la photogrammétrie mais lui sont greffés en amont les traitements vidéo et en aval la conversion des données et la mise en forme des résultats.

## 3 Méthodologie

**Découpage de la vidéo**

MicMac n'a pas été conçu pour le traitement de vidéos. Il possède des outils qui permettent de se replacer en configuration images. Ces outils sont dans un premier temps éprouvés sur des cas variés de vol : drone, parapente... Cela permet de comprendre l'influence du paramétrage de la fonction DIV. Celui-ci peut être résumé aux trois arguments suivants exposés dans l'ordre de prépondérance pour le filtrage des images :

- le pourcentage d'images supprimées d'office quelle que soit la fréquence FPS[3] de la vidéo
- le taux de recouvrement qui est fixé par défaut à 80% mais qui peut être modulé afin de décrire au mieux la situation de vol (par exemple pour une vidéo très résolue et avec une bonne visibilité du sol, 50% de recouvrement peut suffire)
- le nombre d'images à la seconde qui est plus théorique puisque si le recouvrement cible précédemment expliqué n'est pas atteint avec ce ratio, MicMac densifie automatiquement.

Il faut noter que l'algorithme ne conserve que les images les plus à même de fournir des points d'intérêts, c'est-à-dire les images nettes et contrastées [8]. Le contraste est évalué par analyse des histogrammes RVB (i.e. rouge, vert, bleu). Quant à la netteté, aucun indicateur ne la donne directement. Le postulat du raisonnement est le fait que pour une vidéo dont la fréquence FPS est élevée[4], deux images successives ont quasiment la même emprise spatiale. Il est possible de calculer leur corrélation croisée qui s'apparente alors à un critère d'autocorrélation et permet d'identifier une image qui est nette par rapport à son voisinage. En effet si une image est nette, elle contient des détails haute-fréquence qui autocorrèlent mal avec le voisinage. En revanche si elle est floue, les hautes fréquences sont lissées et l'image autocorrèle bien.

La répartition des images selon un ratio est donc modulée par le critère de netteté. La vidéo est découpée en intervalles réguliers selon le nombre d'images par seconde souhaité, puis autour de chaque borne les images voisines sont testées pour retenir l'image la plus nette.

**Calibration des images**

L'étape de calibration est consécutive à ce prétraitement et requiert de connaître parfaitement la caméra. Or à la suite d'un accident d'avion, elle est la plupart du temps détruite ou endommagée de sorte qu'une calibration *a posteriori* est inenvisageable. Il faut donc procéder par approximation. Si le modèle de caméra est connu et pour un accident qui justifie l'investissement, le modèle est racheté et étalonné selon différents protocoles. Le premier est le calcul de la focale par mesure de l'angle de champ (noté FOV[5]) diagonal (exprimé en radians). Dans le cas idéal d'une lentille sans distorsion, la trigonométrie montre que :

$$focale = \frac{diag}{2\tan\left(\frac{FOV}{2}\right)} \quad (1)$$

Mais la plupart du temps, ce modèle n'est pas valable. Dans le cas de lentilles fish-eye comme avec les GoPro, les deux modèles les plus appropriés sont les :

- projection équidistante (ou azimutale équidistante)

$$focale = \frac{diag}{FOV} \quad (2)$$

- projection équisolide (ou azimutale équivalente)

$$focale = \frac{diag}{4\sin\left(\frac{FOV}{4}\right)} \quad (3)$$

Ces projections ont été choisies empiriquement par comparaison avec les valeurs réelles de focales comme le prouve le tableau suivant :

| Focale | réelle | équidistante | équisolide |
|---|---|---|---|
| HERO3 Silver | 2,5 | **2,56** | 2,69 |
| HERO3+ Black | 2,8 | **2,83** | 2,97 |
| HERO5 Black | 3 | 2,87 | **3,02** |
| HERO5 Session | 2,2 | 2,10 | **2,20** |

Cette précision de l'ordre du dixième de millimètre est suffisante pour initialiser un calcul MicMac. La mesure de l'angle de champ repose sur un outil développé au sein du BEA : un double rapporteur gradué.

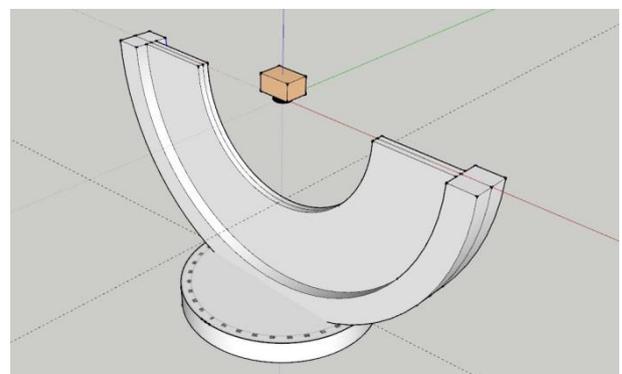

Figure 1 – Double rapporteur

---

[3] *Frames Per Second* : images par seconde
[4] Critères PAL/SÉCAM de 25 images/seconde pour la France et l'Europe, NTSC de 30 images/seconde pour les USA
[5] *Field of view*

Le fonctionnement de ce double rapporteur est fondé sur l'alignement des graduations des deux rapporteurs :

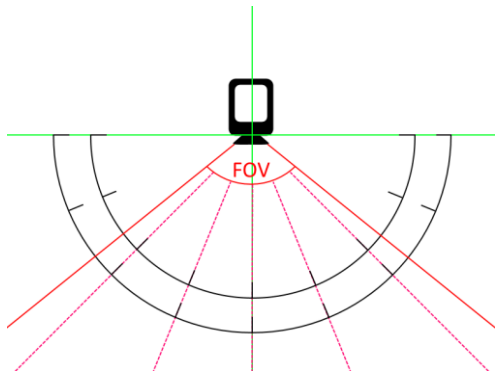

Figure 2 – Alignement de la caméra

Le centre optique de la caméra (à l'intersection des deux traits verts) est alors aligné horizontalement et verticalement (faisceaux magenta) avec le rapporteur et la lecture angulaire faite sur celui-ci donne la valeur de l'angle de champ.

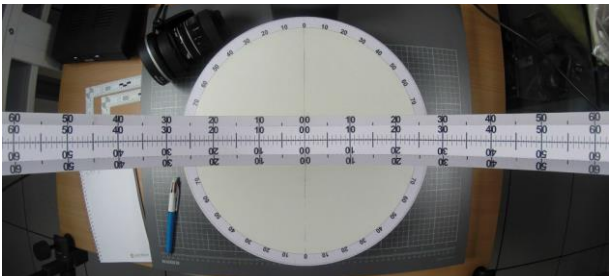

Figure 3 – Mesure de l'angle de champ

En alignant la diagonale de l'image sur l'axe transversal du rapporteur, on lit l'angle de champ diagonal, en l'alignant horizontalement (comme sur la Figure 3), on obtient l'angle horizontal, etc. L'outil donne un angle de champ réel et non théorique puisque l'intégralité des photosites n'est pas nécessairement utilisée.

**Géoréférencement de la caméra**

Une fois les paramètres de la caméra estimés et les images calibrées, il faut géoréférencer la mise en place relative issue de cette calibration [9]. En reprenant les cas de figure présentés en introduction, l'utilisateur peut soit relever une caméra mobile ou positionner un objet mobile sur une vidéo (avec caméra fixe ou mobile). Le relèvement est la forme classique d'un chantier sous MicMac avec des paires d'images en recouvrement. En revanche la localisation d'un objet sur des images est plus délicate. Que la caméra soit fixe ou mobile, il faut connaître sa position à tout instant. Si elle est fixe, ne disposant pas d'autre moyen de calibration, la fonction `Init11P` permet d'orienter les clichés en proposant un auto-étalonnage par résolution de la formule d'image avec des pointés manuels. S'ensuit le positionnement « en relatif » des objets visibles sur la vidéo. La fonction `AlphaGet27` est développée pour cette opération et intégrée dans les dépôts de MicMac.

Cette fonction permet, à partir des orientations de la caméra, ainsi que de plans quadrillés (de profil, en face et de dessus) de l'objet avec ses dimensions réelles, de relever la caméra par rapport à l'objet.

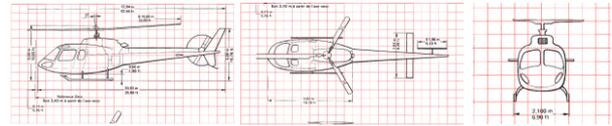

Ces plans permettent de choisir des points dans le référentiel de l'aéronef (un hélicoptère en l'occurrence) qui sont pointables sur les images. Les coordonnées sont exprimées dans un repère local lié au centre de gravité de l'aéronef. Dans cet exemple les points caractéristiques visibles sur la vidéo[6] sont l'axe du rotor, les extrémités des pales, le rotor anticouple (RAC), le nez, les patins, etc. Au total 7 points peuvent être saisis avec la fonction. Or pour un relèvement, 4 points suffisent (3 + 1 de contrôle). La méthode semble donc cohérente. Ainsi les coordonnées des sommets de prise de vue sont connues dans un repère métrique centré sur l'objet, donc la distance entre les deux est également connue à chaque instant.

La fonction sert également à corriger la direction de faisceau d'un pointé image passant par le centre de gravité de l'objet. Cette correction utilise la calibration préalable de la caméra pour appliquer le polynôme de distorsion inverse au faisceau. En normant le vecteur issu de ce faisceau, la fonction renvoie la direction 3D entre la caméra et l'objet.

En multipliant distance et vecteur directeur entre caméra et objet, il en résulte une translation 3D à appliquer aux coordonnées de la caméra pour trouver les coordonnées géographiques de l'objet à l'instant i :

$$\overrightarrow{t_{cam,i}^{obj}} = d_{cam,i}^{obj} \times \overrightarrow{v_{cam,i}^{obj}} \qquad (4)$$

$$\begin{pmatrix} x \\ y \\ z \end{pmatrix}_{obj}^{i} = \begin{pmatrix} x \\ y \\ z \end{pmatrix}_{cam}^{i} + \overrightarrow{t_{cam,i}^{obj}} \qquad (5)$$

Le schema suivant illustre le positionnement d'une caméra à bord de l'avion rouge avec l'étape de relèvement sur objet tierce (avion bleu).

---

[6] La vidéo de l'accident concerne une enquête en cours et les images extraites sont soumises à la confidentialité.

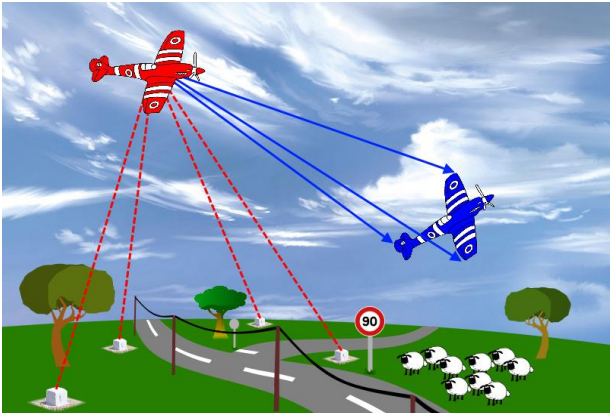

Figure 4 – Positionnement d'un objet en relatif

**Conversion des résultats**

Une dernière étape subsiste pour pouvoir exploiter pleinement les résultats. MicMac calcule des matrices d'orientations avec des angles photogrammétriques notés traditionnellement $(\omega, \phi, \kappa)$. Plusieurs conventions existent pour ce repère et celle adoptée par MicMac est décrite dans sa documentation. Or pour pouvoir reconstituer les mouvements d'un avion, le BEA a besoin des tangage, roulis et lacet notés respectivement $(\theta, \varphi, \psi)$. La conversion entre les deux repères n'est pas triviale et la matrice de passage est la suivante :

$$R_{photo}^{nav} = \begin{pmatrix} \cos(\phi)\cos(\kappa) \\ \cos(\omega)\sin(\kappa)+\sin(\omega)\sin(\phi)\cos(\kappa) & \cdots \\ \sin(\omega)\sin(\kappa)-\cos(\omega)\sin(\phi)\cos(\kappa) \\ \cos(\phi)\sin(\kappa) & -\sin(\phi) \\ -\cos(\omega)\cos(\kappa)+\sin(\omega)\sin(\phi)\sin(\kappa) & \sin(\omega)\cos(\phi) \\ -\sin(\omega)\cos(\kappa)-\cos(\omega)\sin(\phi)\sin(\kappa) & -\cos(\omega)\cos(\phi) \end{pmatrix} \quad (6)$$

Elle correspond au produit de trois rotations selon les axes des x, y puis z. Une fois cette rotation 3D appliquée, il reste à déterminer la rotation entre le repère caméra et le repère avion. Si la caméra est rigidement liée à ce dernier et que son orientation au moment de l'accident est connue, une mesure sur trois axes d'écarts angulaires permet de composer cette matrice. En résumé pour obtenir les angles de navigation de l'avion, il faut exécuter la séquence de transformations suivante [12] :

$$\begin{pmatrix} \theta \\ \varphi \\ \psi \end{pmatrix}_{avion} = R_{cam}^{avion} * R_{photo}^{nav} * \begin{pmatrix} \omega \\ \phi \\ \kappa \end{pmatrix}_{photo} \quad (7)$$

L'export aux formats CSV ou KML est alors possible pour visualisation sous Google Earth ou sous forme d'animation 3D avec le moteur Cesium du BEA.

## 4 Résultats

### Vol d'essai

Le BEA sera amené, avec l'utilisation de MicMac pour l'établissement de ses trajectographies 4D, à citer la librairie photogrammétrique open-source de l'ENSG. Les travaux réalisés sont des éléments factuels importants sur lesquels une grande partie de l'analyse s'appuiera. Les résultats doivent être expliqués et leur validité démontrée. L'enjeu de la justification et de la validation du procédé MicMac est donc décisif pour que le BEA garde sa crédibilité auprès des autorités avec lesquelles il travaille.

C'est pourquoi il a été décidé dans le cadre de mon stage d'organiser une campagne de vols tests reproduisant les conditions réelles d'accident mais en ayant un contrôle sur les événements afin de pouvoir mesurer et comparer toutes les étapes du traitement photogrammétrique.

Pour ce faire, le BEA a fait appel à un prestataire spécialisé dans le vol de drones. Sur la base aérienne militaire désaffectée de Brétigny-sur-Orge, la société Drones-center[7] a mis à disposition un drone Ranger EX emportant à chaque tour deux GoPro, une en oblique avant et l'autre à la verticale. Le terrain de survol a été préalablement balisé avec des cibles lasergrammétriques géantes (2m30 de diamètre) qui servent de points d'appui lors de l'aéro-triangulation :

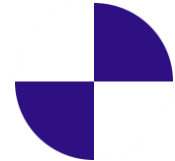

Figure 5 – Schema d'une cible lasergrammétrique

Le drone évolue 100 mètres au-dessus du sol et suit une trajectoire préprogrammée afin de tester des virages, descentes, montées plus ou moins marqués rencontrés dans les cas réels. Si bien qu'avec de telles cibles, sur une vidéo en 720p d'une GoPro dont la largeur du capteur mesure 4,3 mm et la focale 3 mm, elles remplissent environ 12 pixels de l'image.

$$T_{cible}^{image,px} = T_{cible}^{réel} \times \frac{3 \cdot 10^{-3}}{100} \times \frac{720}{4,3 \cdot 10^{-3}} \approx 11,6 \text{ px} \quad (8)$$

---

[7] http://www.drones-center.com/

Des images des vols figurent dans la mosaïque ci-après :

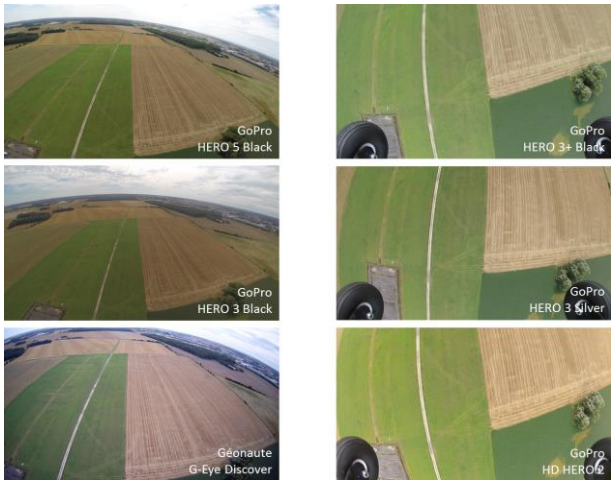

Figure 6 – Images comparatives des 6 caméras embarquées
De haut en bas :
GoPro HERO5 Black – GoPro HERO3+ Black (4K)
GoPro HERO3 Black – GoPro HERO3 Silver (1080p)
Geonaute G-Eye Discover – GoPro HD HERO2 (720p)

Les trajectoires calculées pour chaque couple de caméras sont comparées avec la trace GPS du drone.

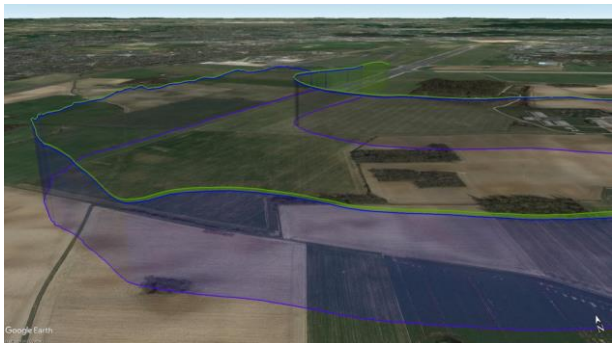

Figure 7 – Différence entre GPS et calcul sur le vol d'essai

■ Trajectoire calculée   ■ Trace GPS

Ci-dessus figure le comparatif réalisé à partir du calcul sur la vidéo 4K enregistrée par la GoPro HERO5 Black. Après analyse des logs de vols (ce sont les données enregistrées par le drone) et de MicMac, l'écart de position moyen entre le GPS et MicMac vaut 1 mètre en planimétrie. L'écart maximal atteint 2,7 mètres. En altimétrie, un biais relativement constant inférieur à 5 mètres est observé. Il faut garder à l'esprit que la trace GPS n'est pas la position vraie et que le GPS a une précision propre (qui est d'ailleurs généralement dégradée en altimétrie). Les vidéos d'autres résolutions (Full HD et HD) donnent des résultats contrastés : la précision trajectographique des vidéos en 1080p rentre dans les seuils de tolérance de 5 mètres dans le plan et la dizaine de mètres en altitude.

L'influence de la qualité des images est aussi testée. Disposant de trois couples de caméras, le premier filmait en 4K UHD, le deuxième en Full HD 1080p et le troisième en HD 720p. Le comparatif porte sur le nombre de points d'intérêts détectés sur des images ayant la même emprise (par ordre décroissant de résolution) :

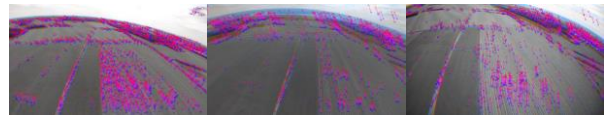

Figure 8 – Nombre de points d'intérêt selon la résolution

HERO5 Black     HERO3 Black     G-Eye Geonaute

Un fort écart est observé entre la 4K et les deux autres qualités d'image. En revanche 1080p et 720p ne se démarquent pas. L'apport de la 4K est donc indéniable au vu de ce bilan. Il faut mentionner que les conditions d'acquisition ne sont pas totalement similaires puisqu'entre le premier et le troisième vol, la luminosité ambiante a changé.

Le positionnement en relatif a été restitué lors de ces essais par l'intermédiaire d'un drone DJI Phantom 4 filmant en 4K le décollage et l'atterrissage du drone porteur des GoPro. Les KML comparés donnent :

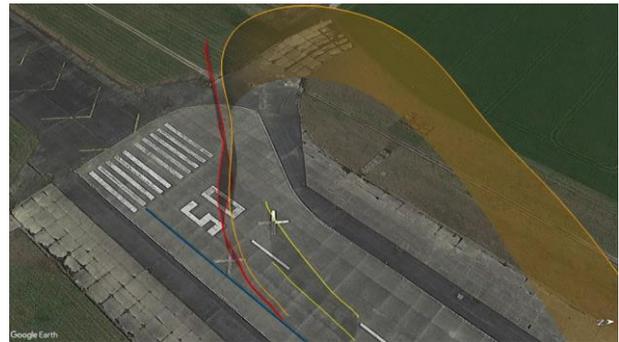

Figure 9 – Différence entre GPS et calcul « en relatif »

■ Trajectoire AlphaGet27   ■ Trace GPS   ■ Trace Phantom 4

Un décalage planimétrique est observé mais l'analyse d'image (avec la ligne de dalle au sol marquée en bleu) montre que le récepteur GPS positionne trop à droite :

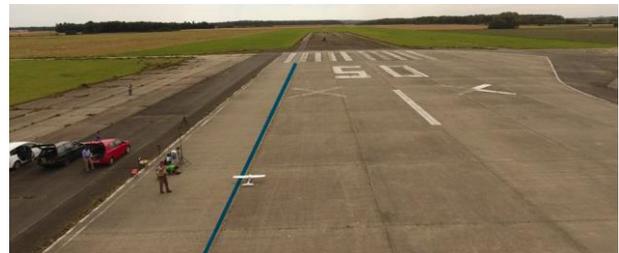

Figure 10 – Repère terrain pour validation de la procédure

Six vidéos d'accidents différents ont été analysées et trajectographiées. Les aéronefs impliqués sont trois avions de voltige, une montgolfière, un avion de chasse, un hélicoptère, un autogire et un parachutiste. Le choix des accidents traités respecte une logique de diversification des configurations de vol afin de tester la robus-tesse et l'adaptabilité de l'outil. Le bilan global des traitements effectués est le suivant : quatre cas sur six ont donné des résultats directement exploitables par les enquêteurs du BEA. Les deux cas restants concernent un accident d'avion de voltige qui a permis d'identifier une limite au niveau des distorsions induites par la traversée de la verrière ; et une collision entre un parachute et un autogire filmée par un parachutiste tiers.

**Limites au calcul**

Ces deux dernières enquêtes permettent de mettre en lumière des limites inhérentes aux données sources issues de vidéos embarquées. La première étude teste la flexibilité de la calibration par MicMac. Pour resituer le cadre de l'accident : la GoPro est posée sur la plage avant du cockpit et filme le pilote ainsi que la partie arrière de la verrière. L'angle de champ vaut 90°. La conséquence d'une telle orientation de caméra est une forte distorsion induite par la traversée de verrière. En effet les lois de l'optique géométrique (lois de Snell-Descartes) rappellent que les faisceaux lumineux sont déviés à l'interface de deux milieux d'indices de réfraction différents. Or les rayons lumineux n'incident pas la vitre avec le même angle de par la forme de goutte d'eau de la verrière.

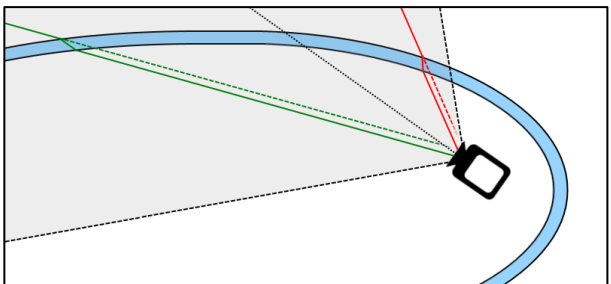

Figure 11 – Schema de la distorsion induite par la verrière

Sur ce dessin, les rayons vert et rouge sont de part et d'autre de l'axe optique mais subissent une déviation tous les deux dans le sens horaire. Cette asymétrie dans la déviation engendre des déformations non modélisables par de simples polynômes de degrés impairs. La seule issue à envisager est de reconstituer le cockpit et de replacer la caméra en l'état au moment de l'accident et de s'étalonner par exemple sur une grille ou un polygone de calibration à l'extérieur de la verrière.

Au final la caméra n'étant pas dans l'axe de l'avion, cette déviation n'est donc aucunement symétrique et vient s'ajouter à la convergence de la lentille. Cependant MicMac s'appuie pour la résolution des équations de calibration sur des modèles de distorsion paramétrés pour décrire un nombre fini et représentatif de possibilités (radial, fish-eye, symétrique, avec décentrement, etc.). Ici la distorsion ne rentre pas dans ces critères. Le processus est donc mis en échec dès la calibration.

Dans le second accident problématique, la méthodologie employée est celle du positionnement en relatif. Concrètement la caméra filmant la scène est portée par un parachutiste au niveau de la tête et l'aplomb du parachute et de l'autogire impliqués. L'étape préliminaire consiste donc en le géopositionnement de la caméra. Or la descente d'un parachutiste est soumise aux contraintes éoliennes qui entraînent des changements brusques d'orientations de la caméra (ainsi que les mouvements du parachutiste lui-même). Ce problème a été identifié lors des tirs à blanc sur une GoPro fixée sur un parapentiste. La trajectoire en elle-même est exploitable bien que son utilisation pour positionner relativement un autre objet est plus complexe.

En outre l'autre face du problème est la taille insuffisante des objets à positionner sur les images. Comme énoncé dans la partie « Méthodologie », la condition d'application d'`AlphaGet27` est entre autres que les objets vus soient assez grands pour pouvoir pointer des détails précisément. Ici l'autogire n'est perçu que par ses pales (au total 3 points dont les deux extrémités et le rotor central) et le parachute par les quatre coins de sa voilure. C'est théoriquement suffisant pour un relèvement mais le manque de redondance (i.e. de contrôle) provoque une trajectographie fausse (surtout pour l'autogire). Cela a permis la détermination d'un seuil minimal de taille en dessous duquel la méthode échoue. Ce seuil vaut environ 20 px de diagonale et 200 px² de surface pour décrire à la fois la distance entre les points et leurs répartitions relatives.

**Cas d'accidents traités**

Parmi les quatre enquêtes dont l'analyse des vidéos a donné des résultats probants, deux trajectoires ont servi à l'établissement de conclusions sur les circonstances de l'accident et sont incluses dans les rapports publiés. Cela met en exergue le potentiel de réactivité de l'outil à partir de la réception des données multimédias jusqu'à la publication officielle. De plusieurs mois (voire années) de calculs manuels, la trajectographie est passée à quelques jours de traitements automatisés.

La suite du paragraphe se focalise sur le cas où la trajectographie a apporté une analyse déterminante à l'enquête concernant l'atterrissage forcé d'une montgolfière. L'enjeu était d'évaluer les mouvements de la nacelle lors de la phase de descente. Contrairement aux autres types d'aéronefs étudiés jusqu'alors, la montgolfière présente plusieurs particularités consécutives à la faible vitesse de déplacement : une vidéo longue (plus de 7 minutes, soit 13 000 images potentielles), un défilement lent du sol – et donc une fréquence d'images pour le calcul photogrammétrique bien inférieure, une luminosité changeante entre le début et la fin, etc… En ayant pris en compte ces spécificités, la trajectoire a été calculée en deux étapes : en premier lieu la phase de croisière de l'aérostat, puis la phase de descente plus rapide et qui concentre l'attention du BEA. Cette stratégie s'est avérée nécessaire car cette seconde partie de vol donnait lieu dans une première itération à un phénomène anormal : le ballon restait sur place tandis que les points du sol montaient progressivement vers celui-ci. Cet « effet de zoom » inattendu a été court-circuité en figeant une calibration (dont la focale) de la première partie pour mettre en place la descente. Les deux trajectoires ont ensuite été raccordées à l'aide d'un algorithme de fusion de mises en place dont le fonctionnement est décrit par la suite. En laissant au préalable quelques images en commun aux deux fichiers d'orientation, il est en effet possible de recalculer une orientation globale qui va ajuster au mieux celles passées en entrée. Il s'agit ici de compenser localement les décalages de position et d'orientation entre les sommets de prises de vue correspondant à une même photo.

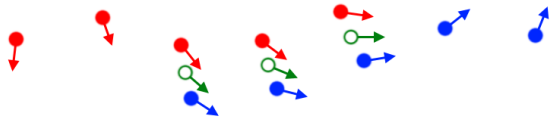

Figure 12 – Illustration de la fusion d'orientations

Cette fonction se nomme Morito dans MicMac. Dès lors l'enquêteur a pu s'appuyer sur des données fiables pour expliquer l'accident, notamment le changement d'orientation de la nacelle (rigidement liée au référentiel caméra) qui est la principale cause. En outre une analyse visuelle des actions du pilote sur les brûleurs a été ajoutée comme métadonnée de trajectoire.

L'autre accident présenté ci-après est un cas d'application du positionnement dit « en relatif ». Il s'agit de la collision entre deux avions de voltige alors que le premier, un Cap 21 (immatriculé F-GLOT), porteur de la caméra, sortait d'une ressource[8] et le second, un Cap 10 (immatriculé F-GUMI), visible sur la partie terminale de la vidéo, entamait un *peel-off* [9]. La vidéo dure 12 secondes durant lesquelles le Cap 10 entre dans le champ de manière fugace :

---
[8] Une ressource est un changement de trajectoire dans le plan vertical pour atteindre un palier.
[9] La manœuvre du « peel-off » consiste à survoler le seuil de piste à très faible hauteur, grande vitesse, et effectuer un virage en montant très cabré et à fort taux de virage.

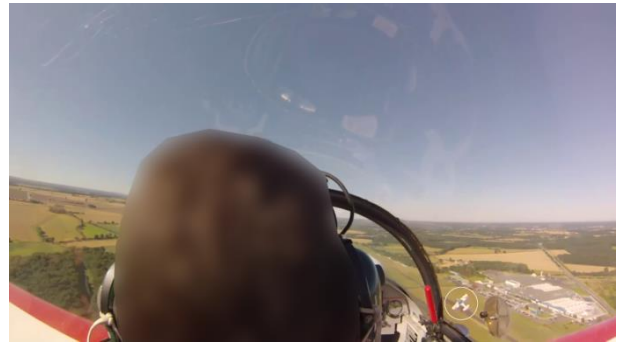

Figure 13 – Dernière image de la vidéo
Le Cap 10 est entouré en blanc dans le coin inférieur droit.

Pour positionner le Cap 10, la méthode `AlphaGet27` est utilisée. MicMac dispose de 7 points caractéristiques (le nez, les saumons d'aile, les coins entre les ailes et le fuselage, les gouvernes de profondeur) et bien définis sur les images. Avec le plan métrique du Mudry Cap 10, les trajectoires suivantes sont établies :

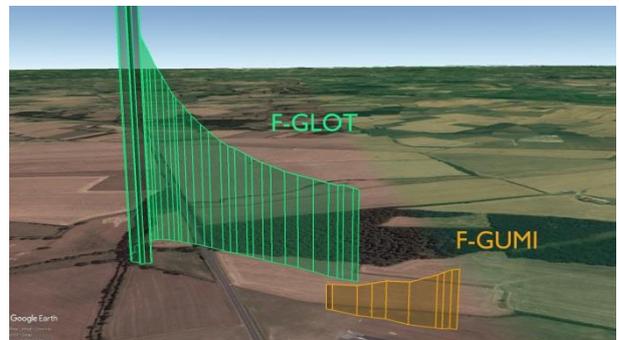

Figure 14 – Vue des deux trajectographies en 3D

En extrapolant d'après les témoignages de personnes ayant assistés à la collision, il est alors possible de resituer le point de contact à moins de 20 mètres près (sphère de confiance de rayon égal à 10 m).

## 5 Conclusions

Le BEA cherchait au travers de cette étude un processus efficace pour générer des trajectoires support de ses enquêtes. L'enjeu n'était donc pas seulement d'expérimenter différentes méthodes mais d'adapter un logiciel libre à la problématique des accidents d'avion tout en le fiabilisant via des tests dans différentes configurations de vol.

Les algorithmes présentés dans cette étude trouvent des applications bien au-delà du cadre du BEA pour lequel ils ont été développés. Ce document démontre en effet que la photogrammétrie ne s'appuie pas uniquement sur des données de prise de vue stéréoscopique et que les traitements sur une simple image apportent des alternatives lorsque la caméra est fixe ou qu'un objet intéressant apparaît dans le champ.

Les méthodes conçues dans l'environnement de Mic-

Mac apportent une approche inédite à la photogrammétrie puisque la caméra se retrouve à la fois comme acteur (elle conduit à la position de l'avion) et spectateur (elle filme l'accident d'un point de vue extérieur). Cette dualité demande une coordination nouvelle pour référencer les aéronefs dans le temps et dans l'espace.

Ces deux axes sont majeurs pour les acteurs du monde de la sécurité des transports routier [11], aérien [12] et spatial [13] qui aujourd'hui doivent pouvoir compter sur des données de géolocalisation précises afin de calculer à leur tour les indices de vitesse instantanée, de facteur de charge, d'accélération, de force centripète, etc. L'automatisation de la chaîne de traitements permettra de produire des animations 3D à partir de vidéos sur tout endroit du monde.

## Références